\documentclass[conference]{IEEEtran}

\usepackage[utf8]{inputenc}
\usepackage[T1]{fontenc}
\usepackage[english]{babel}

\usepackage{blindtext, graphicx}
\usepackage{amsmath}
\usepackage{amssymb}
\usepackage{url}
\usepackage{subcaption}
\usepackage{acronym}
\usepackage{threeparttable}
\usepackage{tabularx}

\usepackage{listings}
\usepackage{parskip}
\usepackage{xspace}
\usepackage{xcolor}
\usepackage[bookmarks=false]{hyperref}
\usepackage{booktabs}
\usepackage{multirow}
\usepackage{listings}
\usepackage[bottom]{footmisc}
\usepackage{algorithm}
\usepackage[noend]{algpseudocode}
\usepackage{enumitem}
\usepackage{verbatim}

\ifCLASSINFOpdf
\usepackage{epstopdf}
\graphicspath{{img/}}
\else
\fi

\acrodef{ann}	[\textsc{\scshape{Ann\xspace}}]		{Artificial Neural Network}
\acrodef{bvlc}  [\textsc{\scshape{Bvlc\xspace}}]	{Berkeley Vision and Learning Center}
\acrodef{cc}	[\textsc{\scshape{CC\xspace}}]		{Connected Component}
\acrodef{cer}	[\textsc{\scshape{Cer\xspace}}]		{Character Error Rate}
\acrodef{cnn}	[\textsc{\scshape{Cnn\xspace}}]		{Convolutional Neural Networks}
\acrodef{ctc}	[\textsc{\scshape{Ctc\xspace}}]		{Connectionist Temporal Classification}
\acrodef{das}	[\textsc{\scshape{Das\xspace}}]		{Document Analysis System}
\acrodef{dip}	[\textsc{\scshape{Dip\xspace}}]		{Document Image Processing Pipeline}
\acrodef{dipp}  [\textsc{\scshape{Dipp\xspace}}]	{Document Image Processing Pipeline}
\acrodef{fcn}	[\textsc{\scshape{Fcn\xspace}}]		{Fully Connected Network}
\acrodef{fp}	[\textsc{\scshape{$FP$\xspace}}]	{False Positives}
\acrodef{fn}	[\textsc{\scshape{$FN$\xspace}}]	{False Negatives}	
\acrodef{gpu}	[\textsc{\scshape{Gpu\xspace}}]		{Graphics Processing Unit}
\acrodef{hmm}	[\textsc{\scshape{Hmm\xspace}}]		{Hidden Markov Model}
\acrodef{htr}	[\textsc{\scshape{HTR\xspace}}]		{Handwritten Text Recognition}
\acrodef{gt}	[\textsc{\scshape{GT\xspace}}]		{Ground Truth}
\acrodef{iu}	[\textsc{\scshape{IU\xspace}}]		{Intersection over Union}
\acrodef{lstm}  [\textsc{\scshape{Lstm\xspace}}]	{Long Short-Term Memory Neural Networks}
\acrodef{pca}	[\textsc{\scshape{Pca\xspace}}]		{Principal Component Analysis}
\acrodef{pkv}	[\textsc{\scshape{Pkv\xspace}}]		{Private Health Insurance}
\acrodef{relu}  [\textsc{ReLUs\xspace}]				{Rectified Linear Units}
\acrodef{rnn}	[\textsc{\scshape{Rnn\xspace}}]		{Recurrent Neural Networks}
\acrodef{svm}	[\textsc{\scshape{Svm\xspace}}]		{Support Vector Machine}
\acrodef{tn}	[\textsc{\scshape{TN\xspace}}]	    {True Negatives}
\acrodef{tp}	[\textsc{\scshape{TP\xspace}}]	    {True Positives}
\acrodef{fn}	[\textsc{\scshape{FN\xspace}}]	    {False Negatives}
\acrodef{fp}	[\textsc{\scshape{FP\xspace}}]	    {False Positives}

\newcommand*{\divahisdb} {\textsc{DIVA-}HisDB\xspace} 

\newcommand*{\fsdb} {{FamilySearch-}DB\xspace} 
\newcommand*{\fscompname} {{Historical Document Reading Challenge on Large Chinese Structured Family Records}\xspace} 
\newcommand*{\fscompacr} {\textsc{ICDAR 2019 HDRC Chinese}\xspace} 

\colorlet{punct}{red!60!black}
\definecolor{background}{HTML}{EEEEEE}
\definecolor{delim}{RGB}{20,105,176}
\colorlet{numb}{magenta!60!black}

\newcommand\JSONnumbervaluestyle{\color{red}}
\newcommand\JSONstringvaluestyle{\color{blue}}

\newif\ifcolonfoundonthisline

\makeatletter

\lstdefinestyle{json}
{
  showstringspaces    = false,
  keywords            = {false,true},
  alsoletter          = 0123456789.,
  morestring          = [s]{"}{"},
  stringstyle         = \ifcolonfoundonthisline\JSONstringvaluestyle\fi,
  MoreSelectCharTable =%
    \lst@DefSaveDef{`:}\colon@json{\processColon@json},
  basicstyle          = \small,
  keywordstyle        = \small\bfseries,
}

\newcommand\processColon@json{%
  \colon@json%
  \ifnum\lst@mode=\lst@Pmode%
    \global\colonfoundonthislinetrue%
  \fi
}

\lst@AddToHook{Output}{%
  \ifcolonfoundonthisline%
    \ifnum\lst@mode=\lst@Pmode%
      \def\lst@thestyle{\JSONnumbervaluestyle}%
    \fi
  \fi
  \lsthk@DetectKeywords%
}

\lst@AddToHook{EOL}%
  {\global\colonfoundonthislinefalse}

\makeatother
\begin{document}

\title{ICDAR 2019 Historical Document Reading Challenge on Large Structured Chinese Family Records}


\author{
    \IEEEauthorblockN{
    Rajkumar Saini\IEEEauthorrefmark{1},
    Derek Dobson \IEEEauthorrefmark{2},
    Jon Morrey\IEEEauthorrefmark{2},
    Marcus Liwicki\IEEEauthorrefmark{1},
    Foteini Simistira Liwicki\IEEEauthorrefmark{1}
    }
    \IEEEauthorblockA{\IEEEauthorrefmark{1}Machine Learning Group, Luleå University of Technology, Sweden, \{\textit{firstname}\}.\{\textit{lastname}\}@ltu.se}   
    \IEEEauthorblockA{\IEEEauthorrefmark{2}FamilySearch, USA, \{\textit{firstname}\}.\{\textit{lastname}\}@familysearch.org}
}

\maketitle

\IEEEpeerreviewmaketitle
\begin{abstract}
We propose a \fscompname, in short \fscompacr. 
The objective of the proposed competition is to recognize and analyze the layout, and finally detect and recognize the textlines and characters of the large historical document collection containing more than 10 000 pages kindly provided by FamilySearch. 

\end{abstract}

\section{Competition protocol and data}
\label{toc:competition}
We invite all researchers and developers in the field of document layout analysis to register and participate in the new \fscompname .

We propose 3 different tasks for this competition:
\begin{itemize}
    \item \textbf{Task 1} Handwritten Character Recognition on extracted textlines
    \item \textbf{Task 2} Layout Analysis on structured historical document images
    \item \textbf{Task 3} Complete, integrated textline detection and recognition on a large dataset
\end{itemize}

\subsection{Dataset}
\label{toc:competition:dataset}

The dataset is provided by FamilySearch~\footnote{\url{https://www.familysearch.org/}} and consists of the following collections:
\begin{itemize}
\item The test set consists of in total 1.135 images selected from 12 separate books.
\item The training set consists of in total 11.715 images selected from another set of 37 separate  books.
\end{itemize}

\fsdb is a collection of Chinese manuscripts 
that have been chosen regarding the complexity of their layout in semantic structure and font. 
All manuscripts are annotated using Aletheia\cite{aletheiaClausner2011}, an advanced system for accurate and yet cost-effective ground truthing of large amounts of documents.
The annotation of the manuscripts are available in PAGE XML format, a sophisticated XML schema which is component of the PAGE (Page Analysis and Ground truth Elements) Format Framework~\cite{pletschacher2010page}.

\subsection{Task Description}
\label{toc:competition:task}
In this competition, we propose 3 different tasks:
\begin{itemize}
    \item \textbf{Task 1} Handwritten Character Recognition on Extracted Textlines
    \item \textbf{Task 2} Layout Analysis on structured historical document images
    \item \textbf{Task 3} Complete, integrated textline detection and recog-nition on a large dataset
\end{itemize}
\subsubsection{Handwritten Character Recognition on Extracted Textlines} 
The scope of this competition is to recognize (OCR) given extracted textlines and, if possible, to find the segmentation points of the characters.
The advantage of the character competition is that we would be able to generate synthetic historical images, once we have the characters segmented and recognized.
Training data will be also available in PAGE-XML format. \\

We will have at least 100 different characters to be recognized, having at least 20 samples, each. The distribution of characters will be according to a typical distribution, so there are actually some characters having more than one thousand instances and thousands of characters having only a few instances. We plan to map less frequent characters to the class label \emph{unknown}.

\subsubsection{Layout Analysis}
The scope of this competition is to segment the page in different classes by assigning a different pixel value for each class:
There are 2 different annotated classes: 
\\
RGB=0b00...1000=0x000008: text (foreground)\\
RGB=0b00...0001=0x000001: non-text (background)\\
The training data will be available as pixel labeled images. 
To avoid unfair penalties for the boundary regions, we add a value for boundary pixels:\\
RGB=0b10...0000=0x800000: boundary pixel (to be combined with one of the classes, expect background)\\
For example, a boundary text is represented as:\\
boundary+text=0x800008
\\
Mislabeling between the foreground and background in the boundary region will not be penalized in the final evaluation (see Section~\ref{toc:evaluation}).

\subsubsection{Textline recongition}
The scope of this competition is to detect and recognize (OCR) a given texline image. 
The training data will be available also in PAGE-XML format.
The PAGE-XML file will contain the information of the textlines’ location and their corresponding text.

\subsection{Submission Types}
\label{toc:competition:submission}
We allow for three different submission formats, either an executable file (or a bash script), a virtual machine, or even a docker image:
\begin{itemize}
    \item \emph{Executable:}
    \begin{itemize}
        \item All dependencies should be in the same (sub)directory;
        \item Provide a Link for downloading the specific zip file.
    \end{itemize}
    \item \emph{VirtualBox Image:}
    \begin{itemize}
        \item Provide a VirtualBox-Image as download link;
        \item Provide instructions how the method can be executed inside the VirtualBox.
    \end{itemize}
    
    \item \emph{Docker Image:}
    \begin{itemize}
        \item Provide the reference image name as hosted on docker hub (see \url{https://hub.docker.com});
        \item Provide instructions how the method inside the docker image can be executed.
    \end{itemize}
\end{itemize}

\section{Evaluation Tools and Metrics}
\label{toc:evaluation}
\subsection{Task 1: Handwritten Character Recognition on extracted
textlines}

The evaluation of \textbf{Task 1} will be based on the edit distance between two text strings as the minimum number of operations (insertion, deletion, and substitution) needed to transform one into the other. 
More details could be found at: https://web.stanford.edu/class/cs124/lec/med.pdf .

The evaluation tool for this task is written in Python and takes two input arguments: 
\begin{itemize}
    \item \textbf{GT-Folder} the folder where the ground truth text files are stored.
    \item \textbf{Predicted-Folder} the folder where the predicted text files are stored.
\end{itemize}

Usage: 
python evalTask1.py GT-Folder Predicted-Folder

\subsection{Task 2: Layout Analysis on structured historical document
images}
The evaluation of \textbf{Task 2} will be similar as in our previous competition and it is freely available as open source on GitHub\footnote{\url{https://github.com/DIVA-DIA/DIVA_Layout_Analysis_Evaluator}}. 
More information about this evaluation tool can be found in~\cite{michele_2017_open}. 

The evaluation of the layout analysis at pixel level is based on the Intersection over Union (IU) as proposed in~\cite{marti2001using} as ranking metric.
The IU, also known as the Jaccard Index, is defined as:
\begin{equation} \label{equ:evaluation:iu}
IU = \frac{TP}{TP + FP + FN}
\end{equation}
where TP denotes the True Positives, FP the False Positives and FN the False Negatives.

For each page, the IU is computed class-wise (background, text, don't care regions) and then averaged.
The final evaluation of a system is then obtained by averaging the IU of all pages of the dataset.

In order to provide the user a more exhaustive evaluation of prediction quality, the tool outputs several other standard metrics, including F1-score, precision, and recall --- for each class and averaged over the classes. 
Additionally, a human-friendly visualization of the results is provided in form of a output image obtained by overlapping the evaluated prediction with the original image. 
This is useful to get a quick estimation of the results and to detect the area of improvement for the evaluated method.


\subsection{Task 3: Complete, integrated textline detection and recognition on a large dataset}
\label{toc:evaluation:task3}
The evaluation of \textbf{Task 3} will be based on the following metrics:
\begin{itemize}
    \item \textbf{insertedNodes} total nodes inserted to transform the one aligned representation into the other.
    \item \textbf{deletedNodes} total nodes deleted.
\item \textbf{substitutedNodes} total nodes substituted.
\item \textbf{deletedNoinsertedEdgesdes} total edges inserted.
\item \textbf{deletedEdges} total edges deleted.
\item \textbf{totalNodes} total number of nodes.
\item \textbf{totalElements} total number of elements in aligned GT representation without counting  an ending graph edge.
\item \textbf{totalErrors} total errors counted.
\item \textbf{errorRatio} error ratio.
\end{itemize}

The evaluation tool for this task is written in Python and takes two input arguments: 
\begin{itemize}
    \item \textbf{GT-Folder} the folder where the ground truth PAGE XML files are stored.
    \item \textbf{Predicted-Folder} the folder where the predicted PAGE XML files are stored.
\end{itemize}

Usage: 
python evalTask3.py GT-Folder Predicted-Folder

\textbf{Important note}. The predicted XML files must have exact the same schema/structure as the provided ground truth XML files. 
Otherwise, if the predicted XML does not match the schema/structure as the provided ground truth XML file, the results of the corresponding XML file shall not be considered and will be instead counted as error. 
If any ground truth XML file found is invalid, it will not be evaluated (please report if such ground truth files are found). 
Note that the order of the lines are important. 
It should be in the same reading order as the given ground truth XML file.

The winner of this task will get an award price of $1.000$ USD provided by FamilySearch.





\section{Acknowledgements}
\label{toc:acknowledgements}
We would like to thank DIVA Group\footnote{\url{https://diuf.unifr.ch/main/diva/}} of University of Fribourg, Switzerland,  and especially Michele Alberti, for providing us the open source evaluation tool for Task 2 (Layout Analysis on structured historical document images).

\ifCLASSOPTIONcaptionsoff
  \newpage
\fi

\bibliographystyle{IEEEtran}
\bibliography{refs}

\end{document}